\documentclass[letterpaper, 10 pt, conference]{ieeeconf}  % Comment this line out if you need a4paper

\IEEEoverridecommandlockouts                              % This command is only needed if 
\usepackage[belowskip=-10pt,aboveskip=5pt]{caption}
\usepackage{graphicx}
\usepackage{cite}
\usepackage{bm}
\usepackage{booktabs}

\title{\LARGE \bf
Scene Induced Multi-Modal Trajectory Forecasting via Planning
}

\author{Nachiket Deo and Mohan M. Trivedi% <-this % stops a space
\thanks{The authors are with the Laboratory for Intelligent and Safe Automobiles,
University of California, San Diego, CA 92092, USA.
        {\tt\small ndeo@ucsd.edu}, {\tt\small mtrivedi@ucsd.edu} }%
}

\begin{document}

\maketitle
\thispagestyle{empty}
\pagestyle{empty}

%%%%%%%%%%%%%%%%%%%%%%%%%%%%%%%%%%%%%%%%%%%%%%%%%%%%%%%%%%%%%%%%%%%%%%%%%%%%%%%%
\begin{abstract}
We address multi-modal trajectory forecasting of agents in unknown scenes by formulating it as a planning problem. We present an approach consisting of three models; a goal prediction model to identify potential goals of the agent, an inverse reinforcement learning model to plan optimal paths to each goal, and a trajectory generator to obtain future trajectories along the planned paths. Analysis of predictions on the Stanford drone dataset, shows generalizability of our approach to novel scenes.

\end{abstract}

%%%%%%%%%%%%%%%%%%%%%%%%%%%%%%%%%%%%%%%%%%%%%%%%%%%%%%%%%%%%%%%%%%%%%%%%%%%%%%%%
\section{Introduction}
To safely and efficiently navigate through spaces shared with humans, autonomous robots need the ability to forecast human motion. An inherent difficulty in motion forecasting is its multi-modal nature. In a given scene, a human can have one of multiple goals, with multiple paths to each goal. Regression based approaches for motion forecasting tend to suffer from mode collapse, resulting in averaged trajectories that may not conform with the scene.           

Prior works have addressed this challenge by learning one-to-many mappings, from available context such as scene cues and past motion, to multiple future trajectories. This is typically done by sampling conditional generative models \cite{gupta2018social, zhao2019multi, lee2017desire, rhinehart2018r2p2, sadeghian2018sophie, bhattacharyya2018accurate}, or learning mixture models \cite{deo2018multi, cui2018multimodal, deo2018convolutional}. However, the high dimensionality of the output space poses a challenge for such models to generalize to novel scenes, especially since each scene can have paths, goals and decision nodes in various configurations. 

Another set of approaches \cite{ziebart2009planning, kitani2012activity, wulfmeier2016watch, zhang2018integrating} pioneered by Ziebart \textit{et al.} \cite{ziebart2009planning} formulate motion forecasting as a reinforcement learning agent exploring a grid defined over the scene. A reward map for the agent is learned via maximum-entropy inverse reinforcement learning (max-ent IRL) \cite{ziebart2008maximum}. This allows for a more intuitive model for the agent's decision making. Also, since the reward map is learned from local scene cues at each grid cell, it allows for better generalization to novel scenes. However, max-ent IRL approaches suffer from two limitations. They require absorbing goal states in the scene to be known beforehand \cite{ziebart2009planning} or uniformly sampled from the scene \cite{kitani2012activity}. More importantly, they can only provide paths taken by the agent in the grid, without mapping them to times in the future. While this is partly addressed in \cite{ziebart2009planning} by decomposing state visitation frequencies over time steps, using a Gaussian distribution, this does not take into account the agent's dynamics. A fast moving agent would make more progress than a slow moving agent along a planned path, over a fixed prediction horizon.

In this work, we propose a planning based approach that can generalize to novel scenes, while not requiring goal states to be known beforehand, and generating continuous valued trajectories, preserving temporal information. Our approach consists of three models:   
%\vspace{-1mm}
\begin{itemize}
    \item \textit{Goal prediction model:} To determine possible goals of the agent, based on the scene and their past motion 
    \item \textit{Optimal path planner:} To determine paths to each sampled goal using max-ent deep IRL
    \item \textit{Trajectory generator:} To output continuous trajectories over the prediction horizon based on the agent's past motion, and an encoding of the planned paths. 
\end{itemize}

\begin{figure}[t]
\centering
\includegraphics[width=\columnwidth]{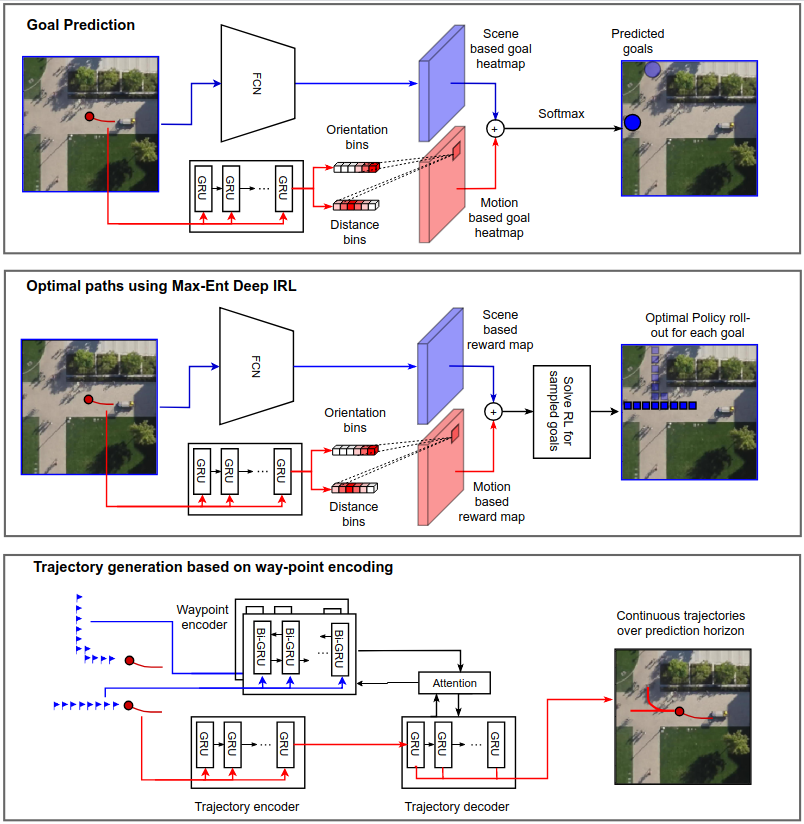}
\caption{Proposed models for scene-induced multimodal trajectory forecasting}
\label{fig_model}
\vspace{-1mm}
\end{figure}

\begin{figure*}[t]
\centering
\includegraphics[width=\textwidth]{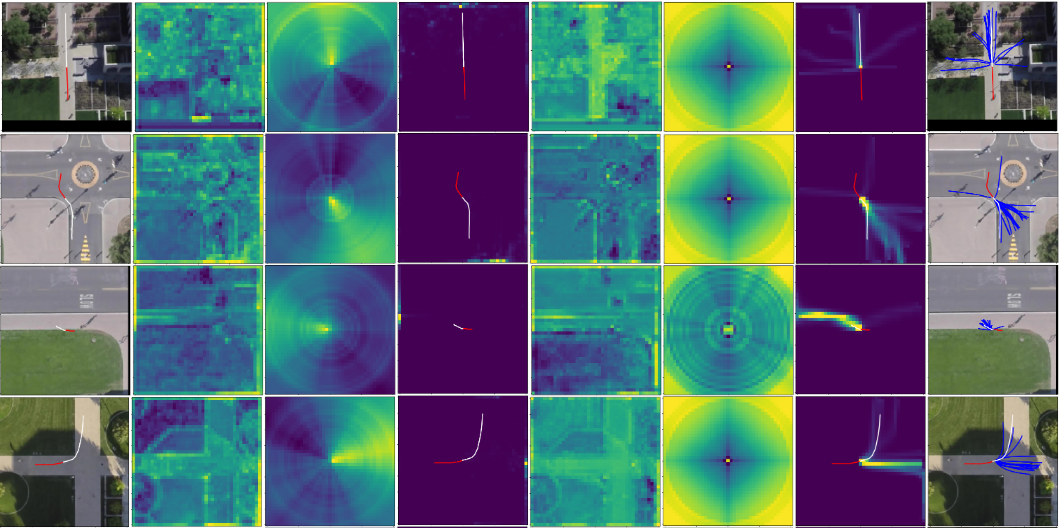}
\caption{\textbf{Example predictions.} From left to right: (1) Scene, past motion(red) and true futre trajectory (white),(2) scene-based goal heatmap, (3) motion based goal heatmap, (4) predicted goals, (5) scene based reward, (6) motion based reward, (7) state visitation frequencies, (8) predicted trajectories (blue) }
\label{fig_examples}
\end{figure*}

\section{Proposed Model}
 
\subsection{Goal Prediction:} The goal prediction model (Fig. \ref{fig_model}, top) consists of two branches; a fully convolutional network (FCN) encoding the scene, and a gated recurrent unit (GRU) encoding the past trajectory. The FCN outputs a heatmap of potential goals in the scene, such as points where paths exit the scene, entrances to buildings etc. The trajectory encoder outputs activations for a discrete set of orientations and distances. These activations are mapped to the 2-D grid by taking a weighted sum of the two nearest distance and orientation values for the center of each cell. The trajectory encoder allows the model to narrow down potential goals based on the past motion of the agent. The two encodings are added and passed through a softmax layer to give goal probabilities on the grid. The model is trained to minimize cross-entropy with respect to the true goal.     

\vspace{-1mm}
\subsection{Optimal paths using Maximum Entropy Deep IRL:} 
We use max-ent deep IRL \cite{wulfmeier2016watch} to learn a reward map on the scene. The reward model (Fig. \ref{fig_model}, middle) is identical in structure to the goal prediction model. However, the heatmaps produced by the FCN and trajectory encoder can be interpreted as the scene-based and motion-based rewards over the grid. Since the FCN processes local patches of the scene, the reward map can generalize to novel scenes. During inference, we solve forward reinforcement learning to get the optimal policy conditioned on goals sampled from the goal prediction model. Roll-outs of the optimal policies give paths to each goal that conform to the scene.  

\vspace{-1mm}
\subsection{Trajectory generation using way-point encoding:}
The trajectory generation model (Fig. \ref{fig_model}, bottom) generates continuous valued trajectories over the prediction horizon, conditioned on past motion and the optimal planned paths. The past trajectory is encoded by a GRU. We treat the grid locations of the planned paths as way-points in the scene. We encode the way-points using a bidirectional GRU encoder. The trajectory generator is a GRU decoder equipped with soft-attention \cite{bahdanau2014neural}. Since all the way-points are typically not reached over the prediction horizon considered, the attention based decoder can attend to relevant way-points as it outputs the trajectory.

\section{Experimental Evaluation:}

\begin{table}[t]
\vspace{0.2in}
\caption{Results on the Stanford drone dataset}
\centering
\begin{tabular}{@{}llllll@{}}
\toprule
 Metric & \begin{tabular}[c]{@{}l@{}}SocialGAN \\ \cite{gupta2018social}\end{tabular} & \begin{tabular}[c]{@{}l@{}}DESIRE*\\ \cite{lee2017desire}\end{tabular} & \begin{tabular}[c]{@{}l@{}}MATF GAN\\ \cite{zhao2019multi}\end{tabular} & \begin{tabular}[c]{@{}l@{}}SoPhie\\ \cite{sadeghian2018sophie}\end{tabular} & Ours  \\ \midrule
mADE            & 27.25                                                                & 19.25                                                             & 22.59                                                              & 16.27                                                            & \textbf{15.73} \\
mFDE            & 41.44                                                                & 34.05                                                             & 33.53                                                              & 29.38                                                            & \textbf{28.18} \\ \bottomrule
\end{tabular}
\label{tab1}
\vspace{0.05in}
*DESIRE uses K=5, while other approaches use K=20
\vspace{-0.2in}
\end{table}

We use the Stanford drone dataset (SDD) \cite{robicquet2016learning} for our experiments. It consists of trajectory data captured at 60 different scenes, with their top-down images. We use the standard benchmark split \cite{sadeghian2018trajnet} for train, validation and test sets. While evaluating a multi-modal trajectory forecasting model, we need to ensure that plausible future trajectories generated by the model that do not correspond to the true future trajectory are not penalized. We thus use the minimum average displacement error (mADE) and minimum final displacement error (mFDE) metrics to evaluate our model:

\begin{equation}
    mADE = \min_{k \in \{1,2, \dots, K\}}\frac{1}{T}\sum_{t=1}^{T}\left \| y_t - \hat{y}^{(k)}_{t}\right \|_{2},
\end{equation}

\begin{equation}
    mFDE = \min_{k \in \{1,2, \dots, K\}}\left \| y_T - \hat{y}^{(k)}_{T}\right \|_{2},
\end{equation}

where $T$ is the prediction horizon, $y_{1:T}$ is the true future trajectory for a given instance, and ${\hat{y}}^{(k)}_{1:T}$ are trajectories sampled from our model. Similar to prior work \cite{gupta2018social, sadeghian2018sophie, zhao2019multi}, we choose $K = 20$, with a prediction horizon of 4.8 s, and past history of 3.2 s.  Table \ref{tab1} shows the mADE and mFDE values for the SDD test set. Our approach achieves state of the art results on SDD. Additionally,
we provide qualitative examples of predictions made by our model, shown in Figure \ref{fig_examples}.  We can observe that our models generate a diverse set of future trajectories that conform with the underlying scene and past motion of the agent.

\addtolength{\textheight}{-12cm}   % This command serves to balance the column lengths

%%%%%%%%%%%%%%%%%%%%%%%%%%%%%%%%%%%%%%%%%%%%%%%%%%%%%%%%%%%%%%%%%%%%%%%%%%%%%%%%

%%%%%%%%%%%%%%%%%%%%%%%%%%%%%%%%%%%%%%%%%%%%%%%%%%%%%%%%%%%%%%%%%%%%%%%%%%%%%%%%

%%%%%%%%%%%%%%%%%%%%%%%%%%%%%%%%%%%%%%%%%%%%%%%%%%%%%%%%%%%%%%%%%%%%%%%%%%%%%%%%

\bibliographystyle{IEEEtran}%IEEEtranS
\bibliography{IEEEexample}

\end{document}